\documentclass{article}
\usepackage{ijcai25}

\usepackage{times}
\usepackage{soul}
\usepackage{url}
\usepackage[hidelinks]{hyperref}
\usepackage[utf8]{inputenc}
\usepackage[small]{caption}
\usepackage{graphicx}
\usepackage{amsmath}
\usepackage{amsthm}
\usepackage{amssymb}
\usepackage{booktabs}
\usepackage{algorithm}
\usepackage{algorithmic}
\usepackage[switch]{lineno}

\newcommand{\ourmodule}{\textsc{mart}}

\title{Leveraging Large Language Models for Active Merchant Non-player Characters}

\author{
Byungjun Kim\textsuperscript{\textdagger}
\and
Minju Kim\textsuperscript{\textdagger}\and
Dayeon Seo\And
Bugeun Kim\textsuperscript{*}\\
\affiliations
Department of Artificial Intelligence, Chung-Ang University, Republic of Korea\\
\emails
\{k36769, minjunim, sdyhappy, bgnkim\}@cau.ac.kr
}

\begin{document}

\maketitle
\def\thefootnote{\textdagger}\footnotetext{Equal contribution.}\def\thefootnote{\arabic{footnote}}

\def\thefootnote{*}\footnotetext{Corresponding author}\def\thefootnote{\arabic{footnote}}

\begin{abstract}
We highlight two significant issues leading to the passivity of current merchant non-player characters (NPCs): \textit{pricing} and \textit{communication}.
While immersive interactions with active NPCs have been a focus, price negotiations between merchant NPCs and players remain underexplored.
First, passive pricing refers to the limited ability of merchants to modify predefined item prices. Second, passive communication means that merchants can only interact with players in a scripted manner.
To tackle these issues and create an active merchant NPC, we propose a merchant framework based on large language models (LLMs), called \ourmodule, which consists of an appraiser module and a negotiator module.
We conducted two experiments to explore various implementation options under different training methods and LLM sizes, considering a range of possible game environments.
Our findings indicate that finetuning methods, such as supervised finetuning (SFT) and knowledge distillation (KD), are effective in using smaller LLMs to implement active merchant NPCs. Additionally, we found three irregular cases arising from the responses of LLMs.
\end{abstract}
\section{Introduction}

Exchanging in-game items is an integral part of open-world role-playing games because the utility of an item depends on the current attributes of a player. For example, a player with high agility may not benefit from an item that enhances agility. Thus, game developers implement an exchange system in their games to enhance gameplay and item utility.

While games commonly feature merchant non-player characters (NPCs) to facilitate item exchanges, these interactions are typically scripted; the merchant only enables players to buy and sell items at fixed prices, but without the dynamics of real-world transactions.
Typically, developers do not allow merchants to alter prices or communicate in free form; instead, they simply present a fixed price to the player without any negotiation. 
However, recent advancements in LLM-integrated games show growing interest in dynamic and flexible interactions, similar to real-world communication where players and NPCs engage in co-constructed dialogue \cite{smoking_gun,aidungeon,unbounded}.
To mirror real-world interactions and enhance player immersion
, we propose a novel framework that leverages LLMs to enable merchant NPCs to engage in actual negotiations with players.

To create an active merchant NPC, which is more aligned to the real world, we need to address two issues that cause the current merchant to act passively: \textit{pricing} and \textit{communication}.
First, regarding passive pricing, the merchant has no authority to adjust item prices. Instead, the game developers decide prices based on the items' utility in the game. However, in the real world, sellers can adjust the prices of their goods according to item specifications.
To the best of our knowledge, gaming industry researchers have not adequately filled this gap.
Since item descriptions often convey sufficient information about utility, we argue that they can be used to estimate the value of unseen items by leveraging information and values from other known items.
Specifically, we let LLMs appraise game items by observing other items.

Second, regarding passive communication, merchants can only interact with players in a scripted manner.
Researchers have explored the adoption of LLMs within games to facilitate more immersive player experiences \cite{goal_oriented_llm,player_driven_emergence}.
However, the way merchants communicate has been given limited focus, which remains a one-way interaction. Current merchant NPCs simply display a predefined list of items for players to choose from, and players respond by clicking on items to make a purchase. This purchasing experience is uniform across all interactions with a merchant, regardless of the individual player involved in the transaction. To enable two-way communication, we propose a negotiation style that mirrors real-world transactions. Through this study, we explore whether LLMs can be used to foster a negotiating interaction within the merchant context.

In this paper, we propose a framework for developing a More Active meRchanT NPC, called \ourmodule. This framework consists of two main components: \textit{appraiser} and \textit{negotiator}. The appraiser module addresses the issue of passive pricing by estimating the value of given items. The negotiator module addresses the passive communication issue by negotiating with players. As game developers have diverse requirements when deploying their games, we conducted experiments to compare potential candidates for each module using a public WoW Classic game item dataset. Specifically, we tested both finetuning methods and $n$-shot prompting methods on Llama 3 models, which come with a wide range of parameter sizes up to 405 billion. Our GitHub repository\footnote{\url{http://github.com/elu-lab/mart}} provides implementation details, prompts, and model outputs.

This study has the following contributions:
\begin{itemize}
\item We propose an LLM-based framework, \ourmodule, for developing active merchant NPCs.
\item Our findings show that LLMs enable merchants to appraise game items, highlighting that supervised finetuning can balance performance, efficiency, and reliability.
\item Our results reveal that LLMs enable merchants to negotiate item prices, highlighting that knowledge distillation efficiently achieves high persuasiveness. 
\item Through statistical and qualitative analyses, we present multiple implementation options for active merchant NPCs, tailored to suit different user preferences.
\end{itemize}

\section{Related Work}


\subsection{Using LLMs to communicate with NPCs}

Recently, game developers have started using LLMs in their games to provide more human-like interaction between players and NPCs. Specifically, there is a growing concern about making conversational NPCs with LLMs \cite{LLM_gamereview,LLM_game_survey_ToG,speechLLMgame,Conversational_Interactions_with_NPCs,chemical,MatyasTheEO}. For example, \citeauthor{speechLLMgame} used LLM-based NPCs to support player interactions within a murder mystery game by helping players interrogate, collect clues, and explore. 
\citeauthor{PANGeA} introduced a framework for integrating LLM-based NPCs with memory to ensure narrative consistency during free-form player interactions, demonstrated in a turn-based, role-playing detective thriller game.

Despite their contributions, prior studies have largely treated LLM-based NPCs as information providers; That is, they only supported simple interactions where NPCs present knowledge or reasoning to the player.
However, more complex forms of interaction that involve mutual influence, such as negotiation, have been largely underexplored in the context of LLM-integrated games.

In real-world applications, NLP researchers have investigated the potential of LLMs to negotiate \cite{persuading_domain,ACE,Nego_assistive}. 
For example, \citeauthor{persuading_domain} leveraged LLMs to generate persuasive dialogues based on everyday scenarios and showed that their approach was more persuasive than other models.
Similarly, \citeauthor{ACE} suggested a personal negotiation coach based on LLMs and proved their effectiveness in assisting human negotiators. 
Inspired by prior work demonstrating the negotiation potential of LLMs, we examine negotiation interactions of an LLM-based NPC in a gaming context.

\subsection{Using LLMs to estimate values}
As item prices are usually fixed in a game system, research has not sufficiently focused on predicting item prices based on item specifications.
An active merchant, however, should have the ability to decide item prices in its shop, similar to a real-world situation.
Thus, we designed an appraiser module, inspired by studies on estimating prices of real assets.

In real-world applications, researchers have investigated machine learning models for estimating commodity prices. Early studies attempted to predict prices based on a fixed set of features \cite{ML_stock,ML_item_price}. For example, \cite{ML_item_price} used a fixed set of features and a machine-learning method for predicting the price of seasonal goods. However, such models generally do not perform well when predicting prices of out-of-domain goods or newly introduced features. To handle such unseen goods or features, researchers utilized latent representations from textual descriptions of items with a language model \cite{LLM_stock,LLM_CTR_predict}. For instance, \citeauthor{LLM_stock} demonstrated that LLMs can predict stock market movements from the earnings report of a company to support information-based investment.
Extending prior works, we demonstrate that LLMs can also achieve strong predictive performance within games.

\section{MART Framework}
\begin{figure*}
    \centering
    \includegraphics[width=0.9\textwidth]{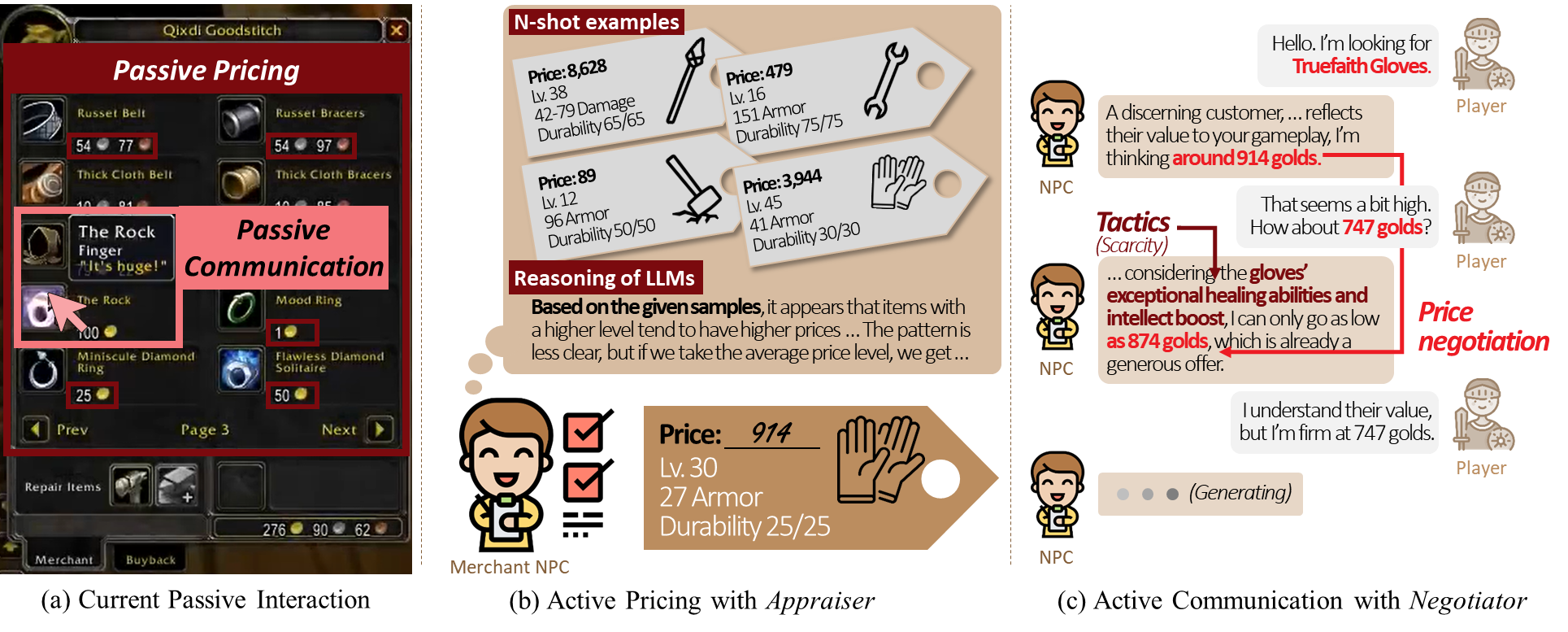}
    \caption{Comparison of current interaction and our \ourmodule\ framework: (a) a screenshot of trader in WoW, borrowed from
    \href{https://www.youtube.com/watch?v=VqVgXp-7h8A&t=706s}{a YouTube video} (b) the proposed Appraiser module, and (c) the proposed Negotiator module.}
    \label{fig:scenario}
\end{figure*}


We propose \ourmodule, an LLM-powered framework for active merchant NPC.
This framework consists of two main modules: \textit{negotiator} and \textit{appraiser}.
Inspired by the behavior of real-world merchants, MART autonomously sets prices for its items and engages in price negotiation with players.
When a player expresses interest in buying an item, the appraiser module suggests a retail price for the item. Based on this suggestion, the negotiator module begins negotiations with players who wish to buy the item from the merchant NPC. This scenario is different from the purchasing experience of players in current games, as shown in Figure \ref{fig:scenario}. 

First, the appraiser module is designed using a language model that interprets an item description to estimate its retail price. The module takes item descriptions presented as natural language sentences, which include details such as required level, effects, and durability.
Since the structure of such descriptions can vary across different games, we adopt a natural language format to ensure versatility.
In Section \ref{sec:appraiser}, we introduce LLM-based appraiser modules using two straightforward approaches and evaluate their performance.

Second, the negotiator module employs a language model to facilitate negotiation dialogues with players. This module receives the item descriptions, retail price, and history of previous conversations. It then generates the appropriate response aimed at persuading the player. Since effective negotiation involves various tactics to persuade players, we experimented LLM-based negotiator modules with two simple approaches that integrate 10 tactics, as detailed in Section \ref{sec:negotiator}.

\section{Appraiser Module}
\label{sec:appraiser}


To develop the appraiser module, we experimented with two approaches: $n$-shot in-context learning (ICL) \cite{icl} and supervised finetuning (SFT).
Here, we set the $n$ as 10 for our ICL approach and called LLMs via public APIs\footnote{\url{http://openrouter.ai}}.
In this section, we outline each approach and illustrate our findings through an experiment on a WoW Classic item dataset.

\subsection{Two tested approaches}

First, to leverage the ability of LLMs to generalize from a set of given examples, we tested ICL method.
ICL enhances reasoning through demonstrations and requires no additional training, making it a simple and efficient method to implement.
Thus, we directly employed LLMs and provided ten random examples in the prompt.
We used Llama models, a popular open-source family of LLMs: Llama-3.1 8b, 405b, and 3.2 1b. Note that ICL does not require any additional training to update LLM parameters.

Second, owing to the substantial computational resources and time required by the 405 billion parameter model, we tested the SFT method on smaller LLMs. To avoid catastrophic forgetting, we froze their parameters and trained an additional adapter. When we input an item description, the smaller LLM transforms the description into a single latent vector. Thereafter, the additional adapter uses the vector to predict the retail price of the item as a regression problem. We used Llama 3.1 8b and 3.2 1b.

\subsection{Used Dataset} 
We used an item dataset from WoW (World of Warcraft) Classic \footnote{\url{https://www.kaggle.com/datasets/mylesoneill/classic-world-of-warcraft-auction-data}} published by Blizzard Entertainment.
By extending the existing WoW Classic item dataset, we created a dataset by crawling item information from the provided URLs for each item and retaining only purchasable items with disclosed prices.
Dataset construction followed three steps. First, we collected 3,376 items by excluding item derivatives from the crawled dataset to simplify the problem.
We then removed 106 items priced below 10 coppers (the cheapest currency unit in WoW) to make enough room for price negotiation during transactions. Finally, we converted all retail prices into coppers, using the conversions of 1 gold to 100 silver and 1 silver to 100 coppers. As a result, we collected a total of 3,270 items, with prices ranging from 10 coppers to 57,018 coppers. More than 50\% of the items were priced below 1,250 coppers, with a median price of 1,238 coppers and an average price of 3,249 coppers. We further divided the dataset into a training set (80\%, 2,616 items), validation set (10\%, 327 items), and test set (10\%, 327 items).

\subsection{Evaluation metrics}

We used four metrics to evaluate the suitability of the two approaches for implementing the appraiser module: \textit{mean absolute percentage error} (MAPE), \textit{standard deviation}, \textit{skewness}, and \textit{unexpected output rate} (UOR).
First, we used MAPE. Due to the wide range of item prices (from 10 to 57,018 coppers), the absolute errors of higher-priced items may overshadow those of lower-priced items. Therefore, we used percentage errors instead of absolute errors. We computed MAPE as the ratio of error to true price, as illustrated in Equation \ref{eq:mape}, where $\hat{y}_i$ is the predicted price and $y^*_i$ is the true price for an item $i$.
\begin{eqnarray}
    PE_i &=& \frac{\hat{y}_i - y^*_i}{y^*_i}\\
    \mathrm{MAPE} &=& \mathbb{E}_i \left[\ | PE_i |\ \right].
    \label{eq:mape}
\end{eqnarray}

Second, we used the standard deviation $\sigma$ of percentage errors ($PE$s). This metric can reveal the extent of variability in appraised prices. If the percentage errors are similar across different items, the standard deviation should be low.

Third, we measured the skewness of $PE$s. Skewness indicates whether a distribution leans toward positive or negative values. Therefore, we believe that analyzing skewness can help us determine if a model tends to underestimate or overestimate item prices. Mathematically, we computed skewness using Equation \ref{eq:skew}, where $\mu$ indicates the mean of $PE$s.
\begin{equation}
    \mathrm{Skewness} = \mathbb{E}_i \left[ \left(\frac{PE_i - \mu}{\sigma}\right)^3 \right].
    \label{eq:skew}
\end{equation}

Lastly, we calculated the rate of unexpected outputs. When LLMs appraise an item, they may generate ambiguous appraisals to extract an exact price from the output sentences. For example, they sometimes suggest multiple prices even if we request a single prediction. We referred to these as unexpected outputs and estimated their frequency among the items. Note that this error only occurred in ICL models because SFT models directly produced the retail price through their prediction head. While LLMs can identify such unexpected cases, we manually labeled these errors.



\subsection{Results}

\begin{table}[!tp]
\begin{center}
\begin{tabular}{llrrrr}
    \toprule
        &  & MAPE & Std. 
        Dev. & Skewness  & UOR \\
    \midrule
    ICL & 1b      & 14.68 & 44.42  & 10.23 & 29.50               \\
        & 8b      & 4.34  & 12.36  & 6.85 & 20.49               \\
        & 405b    & \textbf{1.34}  & 3.34 & \underline{-5.06} & \textbf{5.20}               \\
    \midrule
    SFT  & 1b      & 3.59  & 11.57   & -5.84 & - \\
        & 8b & \underline{2.66} & 11.26 & \textbf{-3.47} & - \\
    \bottomrule
\end{tabular}
\end{center}
\caption{Results of assessing two methods of appraiser}
\label{tab:priceresult}
\end{table}

Table \ref{tab:priceresult} shows the result of our experiments, comparing ICL and SFT methods. Of the five models, the ICL-405b achieved the best performance. It successfully outputted a retail price (94.8\% of the cases) with the lowest MAPE (1.34\%) and produced a few unexpected outputs (5.2\%), while it was slightly skewed toward underestimation (-5.06). The second-best model was the SFT-8b. It exhibited a slightly higher MAPE score (2.66\%) and the lowest absolute skewness toward underestimation (-3.47). Meanwhile, the SFT-1b demonstrated lower performance than the best model (3.59\%) with a similar underestimation (-5.84). These results are different from those of smaller ICL models. The ICL-1b exhibited very high MAPE (14.68\%) with highly overestimated prices (10.23) and a high unexpected output rate (29.05\%). Similarly, the ICL-8b overestimated prices (6.85) with a moderate level of MAPE (4.34) and a high unexpected output rate (20.59\%).

\subsection{Discussion}

We discuss our results in terms of three factors--\textit{performance, efficiency}, and \textit{reliability}--to assess the in-game applicability of the appraiser module.
First, the appraiser has the potential to be used in a game because of its performance. Except for ICL-1b, five models demonstrated a MAPE of less than 5\%. While the difference may seem substantial for high-priced items, it does not exceed 100 coppers for over half of the items, considering that the median price was 1,238 coppers. As gold coins are more frequently used in WoW Classic, a difference of 100 coppers is acceptable. Moreover, when using larger ICL models, the appraised prices aligned more closely with the true prices. Therefore, for game developers seeking a more precise appraiser module, ICL-405b is the best option, as it exhibited an appraisal error of less than 16 coppers for the median price.

Second, the SFT method is efficient for developing an appraiser module. Results show that SFT-8b performed much closer to ICL-405b than to ICL-8b; even the SFT-8b and ICL-8b used almost an identical number of parameters. Also, the SFT-1b outperformed the two ICL models, ICL-1b and ICL-8b. These results imply that it is possible to use a smaller LLM to implement an appraiser within a low-resource environment by finetuning it. Note that the ICL models achieved high performance by observing ten random pairs of item descriptions and prices. Researchers have reported that the performance of ICL methods can be improved by carefully curating examples or increasing the number of examples \cite{icl_demonstration}. However, this curation demands additional human resources, and increasing the number of examples incurs higher computational costs. By contrast, a dataset for SFT can be generated with significantly fewer resources, using pairs of items and their prices.

Third, it is worth noticing that ICL methods can sometimes be unreliable when playing the role of an appraiser. We observed cases in which ICL models produced unexpected outputs. For instance, ICL-405b generated multiple price candidates or a continuous range of prices, despite our request for a single price output. Moreover, smaller ICL models occasionally failed to predict prices. Although we did not adopt any post-processing methods for these outputs, game developers should prevent such failures in their games. Alternatively, developers can use the SFT method, ensuring an LLM predicts a specific price.

\section{Negotiator Module}
\label{sec:negotiator}
We introduce a negotiator module inspired by real-world merchants, aiming to sell items while pursuing profit. 
To develop negotiators, we tested two approaches: zero-shot prompting (ZSP) and knowledge distillation (KD).
We chose different methods from those used in the appraiser module, as supervised negotiation data is difficult to obtain.
In the following subsections, we describe our methods, dataset generation procedure, and details of the experiments. Then, we discuss our findings by comparing these two approaches.

\subsection{Two tested approaches}

First, we employed the ZSP method in which an LLM generates negotiation dialogues without being provided any demonstrations.
We initially considered a na\"ive negotiation method without using any specific tactics. This approach relied solely on using the pretrained knowledge of LLMs, which reflects a general understanding of the world; consequently, the generated negotiation strayed significantly from the intended game setting. Thus, we inputted 10 negotiation tactics within the prompt, as shown in Table \ref{tab:tactics}, which are inspired by \cite{Strategy6,PersuasionGender,ItemkeyValue1,ItemkeyValue2}. We evaluated three Llama variants as in the appraiser. 

\begin{table}
    \centering
    
    \begin{tabular}{p{.25\columnwidth}p{.65\columnwidth}}
        \toprule
        \multicolumn{2}{@{}l}{\textbf{Six persuasion strategies}}\\
        \midrule
        Liking & building relationships through common ground or compliments \\
        \midrule
        Reciprocity & the tendency to return favors \\ 
        \midrule
        Social proof & mimicking observed behaviors \\
        \midrule
        Consistency & aligning with past actions \\ 
        \midrule
        Authority & trusting experts \\ 
        \midrule
        Scarcity & valuing rare items \\
        \toprule
        \multicolumn{2}{@{}l}{\textbf{Four perceived values of a game item}}\\
        \midrule
        Enjoyment & enhancing gaming 
        experience \\ 
        \midrule
        Character competency & leveling up and boosting abilities \\ 
        \midrule
        Visual\quad\quad\quad authority & customizing characters to attract attention \\
        \midrule
        Monetary value & reasonable pricing and good value \\
        \bottomrule
    \end{tabular}
    \caption{Tactics used in the input prompt to help in negotiations}
    \label{tab:tactics}
\end{table}

Second, we employed knowledge distillation to allow smaller LLMs to achieve comparable performance with lower resource requirements.
By transferring the knowledge of a larger model to a smaller model, the distillation reduces computational cost while maintaining performance. We trained student models on a dialogue dataset generated by teacher model.
Here, we prompted the teacher model to select an appropriate negotiation tactic before generating each utterance to distill its persuasiveness.
To further reduce computational demands during distillation, we applied quantization and low-rank adaptation (LoRA) \cite{lora}.
Using these methods, we distilled knowledge about negotiation from Llama 3.1 405b to two smaller LLMs: Llama 3.1 8b and 3.2 1b.

\subsection{Dataset}

We prepared a dataset by generating a negotiation dialogue for each item in the appraiser dataset. To simulate a negotiation dialogue, we used two agents: a merchant and a player. As the merchant, we used Llama 3.1 405B, the teacher model. As the player, we used GPT-4o with input prompts about tactics. We used different models for these two agents to (1) avoid adopting a similar reasoning process and (2) make the player as similar as possible to human beings, as reported in \cite{zeroshotnego1}. To simulate the diversity of real-world customers, we used a temperature value of 1.0 for GPT-4o.

The negotiation procedure was as follows. Before starting a negotiation, both agents received an item description of the negotiation subject as shared information.
The two agents were assigned different prices for each item: the player wanted to buy the item at a discount of 10\% to 25\%.
The player started the negotiation with an utterance ``\texttt{Hello. I'm looking for [the item name]}.''
Although greetings and small talk are common in real-world negotiations, we intentionally avoided including such content to ensure that the models learned to focus on goal-directed negotiation behaviors.

After the initial utterance, the two agents engaged in price negotiation until the player decided whether to purchase the item or not.
Until the decision was made, the merchant kept persuading the player to buy the item to simulate a real-world merchant. The player could terminate the conversation by mentioning ``\texttt{conversation over}.'' To avoid long dialogues, we set the maximum number of turns to 15. As a result, we generated 2,943 conversations: 2,616 conversations from the training set items and 327 conversations from the validation set items. 



\subsection{Evaluation metrics}

We compared three ZSP models and two KD models using three evaluation metrics: \textit{persuasiveness}, \textit{dominance}, and \textit{agreement}. First, we measured the persuasiveness of each utterance generated by the tested negotiators.
Similar to commercial behavior in the real world, the merchant NPC should effectively persuade players to buy or sell an item at a favorable price. Various tactics are used in an effective negotiation; so, we aimed to evaluate the effectiveness of the negotiator in using 10 different tactics to create persuasive statements. To measure persuasiveness, we used G-Eval \cite{geval}, a widely-used evaluation method using LLMs. The method directly asks GPT-4 to evaluate an input text according to given criteria. Specifically, we used a 5-point scale and averaged 20 runs following G-Eval.

Second, we measured the dominance of the merchant over the player during the negotiation. This metric indicates whether the merchant holds more power in the relationship with the player, highlighting the concept of power dynamics \cite{powerdynamics}. 
A human merchant usually has the initiative in price negotiation, rather than giving the initiative to the customer. So, as a negotiated price increases, the merchant profits more while the player incurs greater losses.
In other words, the negotiation is a type of zero-sum game.
Considering such power dynamics, we measured dominance using Equation \ref{eq:dominance}, where $y$, $y^m$, and $y^p$ indicate the agreed price, retail price, and price desired by the player, respectively.
The equation quantifies the ratio between the profit gained by the merchant ($y_i - y_i^p$) and the gap between two agents ($y_i^m- y_i^p$).

\begin{equation}
\mathrm{Dominance} = \mathbb{E}_i \left[ \frac{y_i - y^p_i}{y^m_i - y^p_i} \right]
\label{eq:dominance}
\end{equation}

Third, we defined the agreement rate as the proportion of negotiations in which the merchant and the player reached a mutual settlement on a specific price.
While dominance captures how favorable the final price is to the merchant, relying on it alone may lead to misinterpretation. A merchant who persistently offers unreasonably high prices might achieve high dominance in a few successful cases, while failing most negotiations.
Therefore, we also consider the agreement rate, which measures how often a mutual settlement is reached across total negotiations.


\begin{table}[!tp]
\begin{center}
\begin{tabular}{llr@{ $\pm$ }rr@{ $\pm$ }rr@{\% }}
    \toprule
         &  & \multicolumn{2}{@{}c@{}}{Persuasiveness} & \multicolumn{2}{c}{Dominance} & \multicolumn{1}{c}{Agreement} \\ 
    \midrule
    ZSP  & 1b    & 2.95 & 0.93   & 0.14&0.54 & 90.52\\
         &       & \multicolumn{2}{c}{\scriptsize ($N$=1,650)} & \multicolumn{2}{c}{\scriptsize ($N$=296)} \\
         & 8b    & 3.74&0.63      & \underline{0.42}&0.25 & \textbf{96.94}\\
         &       & \multicolumn{2}{c}{\scriptsize ($N$=1,620)} & \multicolumn{2}{c}{\scriptsize ($N$=317)}\\
         & 405b  & \underline{3.92}&0.53     & \textbf{0.47} & 0.22 & \underline{90.83}\\
         &       & \multicolumn{2}{c}{\scriptsize ($N$=1,502)}  & \multicolumn{2}{c}{\scriptsize ($N$=297)}\\
    \midrule
    KD  & 1b    & 3.65&0.67   & 0.26&0.26  & 82.57\\
        &       & \multicolumn{2}{c}{\scriptsize ($N$=1,359)} & \multicolumn{2}{c}{\scriptsize ($N$=270)}\\
         & 8b    & \textbf{3.99} & 0.50     & 0.40 & 0.17 & 81.04\\
         &       & \multicolumn{2}{c}{\scriptsize ($N$=1,390)} & \multicolumn{2}{c}{\scriptsize ($N$=265)}\\
    \bottomrule
\end{tabular}
\caption{Results of assessing two methods of negotiator. Numbers in parentheses of persuasiveness and dominance indicate the number of utterances and settled negotiations, respectively.}
\label{tab:geval}
\end{center}
\end{table}

\subsection{Results}

Table \ref{tab:geval} shows the results of our experiments, comparing ZSP and KD methods. The results show that the KD-8b performed the best and successfully used persuasive tactics. The model achieved a score of 3.99 in terms of persuasiveness, followed by ZSP-405b (3.92), ZSP-8b (3.74), KD-1b (3.65), and ZSP-1b (2.95). The ZSP-405b had the strongest dominance (0.47), followed by ZSP-8b (0.42), and KD-8b (0.40). The ZSP-8b had the highest agreement (96.94\%), followed by ZSP-405b (90.83\%) and ZSP-1b (90.52\%).

We further performed statistical tests to verify differences between the five models in terms of persuasiveness and dominance. 
We conducted a one-way ANOVA to examine whether there were significant differences among groups using the \texttt{statsmodels} library \cite{statsmodels}. Also, we conducted a post-hoc analysis using the Tukey-HSD test and $p$-value adjustment to identify group differences. 
Tables \ref{tab:persuasiveness} and \ref{tab:dominance} show the statistical result. First, we observed that model differences affect the persuasiveness ($p<$ 0.001). In detail, pairwise differences are all significant: ZSP-405b versus KD-8b ($p$=0.04), ZSP-8b versus KD-1b ($p$=0.004), and the other eight pairs ($p<$ 0.001) are all significant. Second, we also observed that model differences affect the dominance ($p<$ 0.001). Pairwise differences are all significant except for two pairs: ZSP-405b versus ZSP-8b ($p$=0.246) and ZSP-8b versus KD-8b ($p$=0.891) are statistically insignificant.


\newcommand{\starp}{\textsuperscript{*}}
\newcommand{\starpp}{\textsuperscript{**}}
\newcommand{\starppp}{\textsuperscript{***}}

\begin{table}[t]
    \centering
    \begin{tabular}{l@{\quad\textit{vs.}\quad}lrr@{}l}
      \toprule
      \multicolumn{5}{c}{One-way ANOVA: \quad $F_{4,7516}=600.67$, $p<0.001$\starppp}\\
      \midrule
      \multicolumn{2}{c}{Posthoc comparison} & Mean Diff. & Adj. $p$ & \\
      \midrule
      ZSP-1b & ZSP-8b   & 0.789 & $<$0.001 & \starppp \\
      ZSP-1b & ZSP-405b & 0.973& $<$0.001 & \starppp \\
      ZSP-1b & KD-1b    & 0.702& $<$0.001 & \starppp \\
      ZSP-1b & KD-8b    & 1.044& $<$0.001 & \starppp \\
      \midrule
      ZSP-8b & ZSP-405b & 0.184 & $<$0.001 & \starppp\\
      ZSP-8b & KD-1b    & -0.086 & 0.005 & \starpp \\
      ZSP-8b & KD-8b    & 0.255 & $<$0.001 & \starppp \\
      \midrule
      ZSP-405b & KD-1b  & -0.271 & $<$0.001 & \starppp \\
      ZSP-405b & KD-8b  & 0.070 & 0.040 & \starpp \\
      \midrule
      KD-1b & KD-8b     & 0.341 & $<$0.001 & \starppp \\
      \bottomrule
      \multicolumn{5}{r}{\small \starp $p<0.05$, \starpp $p<0.01$, \starppp $p<0.001$}
    \end{tabular}
    \caption{Results of statistical test on persuasiveness}
    \label{tab:persuasiveness}
\end{table}

\begin{table}
    \centering
    
    \begin{tabular}{l@{\quad\textit{vs.}\quad}lrr@{}l}
      \toprule
      \multicolumn{5}{c}{One-way ANOVA: \quad $F_{4,1440}=53.53$, $p<0.001$\starppp}\\
      \midrule
      \multicolumn{2}{c}{Posthoc comparison} & Mean Diff. & Adj. $p$ & \\
      \midrule
      ZSP-1b & ZSP-8b   & 0.279 &  $<$0.001& \starppp \\
      ZSP-1b & ZSP-405b & 0.331 &  $<$0.001& \starppp \\
      ZSP-1b & KD-1b    & 0.118 &  $<$0.001 & \starppp \\
      ZSP-1b & KD-8b    & 0.254 &  $<$0.001 & \starppp \\
      \midrule
      ZSP-8b & ZSP-405b &  0.052 & 0.247 &  \\
      ZSP-8b & KD-1b    & -0.160  &  $<$0.001 & \starppp \\
      ZSP-8b & KD-8b    & -0.024 & 0.891 & \\
      \midrule
      ZSP-405b & KD-1b  & -0.212 & $<$0.001  & \starppp  \\
      ZSP-405b & KD-8b  & -0.076 & 0.036 &  \starp \\
      \midrule
      KD-1b & KD-8b     & 0.136 & $<$0.001 & \starppp \\
      \bottomrule
      \multicolumn{5}{r}{\small \starp $p<0.05$, \starpp $p<0.01$, \starppp $p<0.001$}
    \end{tabular}
    \caption{Results of statistical test on dominance}
    \label{tab:dominance}
\end{table}

\subsection{Discussion}

In open-world games, we believe that merchant NPCs should contribute to immersive player experiences. To support such immersion, developers need to consider two essential parts of a merchant: its \textit{purpose} and \textit{position} within the game world. We discuss our findings with respect to these two aspects.

First, small models can effectively support the objectives of a merchant by using the KD method. Note that the persuasiveness metric measures whether the merchant successfully employed tactics to persuade the player. Therefore, a higher persuasiveness score indicates that the negotiator is better equipped to fulfill the purpose of the merchant. The results show that the largest model (ZSP-405b) achieved a significantly high score. Additionally, we observed that the KD models effectively learnt persuasive tactics from their teacher model, Llama 3.1 405b. In other words, the KD models generally outperformed the ZSP models with the same size. So, it is reasonable to use smaller KD models instead of large ZSP models; we recommend KD-8b as the best option.

Second, the ZSP and KD methods can introduce different merchant personalities. In WoW Classic, there are many merchants in a variety of situations, including differences in race and faction. These factors should be considered when developing merchants to enhance the immersive experience of players. For example, a merchant can use different policies for different players based on their races or factions; a merchant can impose higher prices on rich players or players unfavorable to the merchant. As an early stage, our results imply such distinction is possible using different models. To create an agent favorable to players, developers can opt for models with high agreement rates and low dominance such as ZSP-1b or ZSP-8b. Conversely, if developers are looking for more stubborn merchants, models with low agreement rates and high dominance, like KD-8b, would be a good option.

\subsection{Additional case analysis}

We further analyzed negotiation utterances to identify irregular cases that should be considered when deploying \ourmodule\ in a game. Three of the authors manually labeled utterances of five models and discovered three irregular cases: \textit{giveaways}, \textit{improvisations}, and \textit{arithmetic errors}.

The giveaway case refers to situations where a merchant offers additional items or enhancements to persuade a player to make a purchase. The quote following this paragraph provides an example. We suspect this case occurred because the negotiator used the reciprocity tactic. Although this practice is common in the real world, giveaways are typically not incorporated into a game. Thus, developers should handle this case when integrating negotiators into their game world. If developers allow merchants to provide giveaways, their game system should support such cases. Otherwise, developers should prevent the negotiator from using such a tactic.

\begin{quote}
\ttfamily\small
    \begin{description}
        \item[Player:] \textit{Hello. I'm looking for a Cadet Belt.}
        \item[ZSP-405b:] \textit{You've got a great eye for quality gear, ...}
        \item[]... (omitted) ...
        \item[ZSP-405b:] \textit{I understand your position, but ..., I'll throw in one final sweetener. \textbf{I'll give you a free upgrade} to a premium pouch, ...}
    \end{description}
\end{quote}

The improvisation case indicates that merchants sometimes create spontaneous statements about their stock lists. This error mainly occurred in ZSP-1b, the smallest model without training. An example of this case is shown in the quote following this paragraph. We suspect that this improvisation is a form of ``hallucination.'' In natural language processing, this term describes instances where language models generate nonsensical or inaccurate text in relation to the provided input sources \cite{hallucination}. Prior studies have also shown that smaller language models are more likely to generate hallucinated outputs \cite{hallucination_size}. Therefore, developers should be prepared for the possibility that merchants may suggest nonexistent items. This case can be handled through two approaches: mitigation or approval. To mitigate the problem, we recommend that developers implement a retrieval-augmented generation system \cite{rag}, which encourages language models to produce outputs based on a specific database.
For approval, developers should establish a confirmation system to assess the relevance and legitimacy of any newly introduced items by merchants.
\begin{quote}
\ttfamily
\small
    [The Conjurer's Bracers is in WoW Classic, but the Conjurer's Sigil Cloak is not.]
    \begin{description}
        \item[Player:] \textit{Hello. I'm looking for conjurer's bracers.}
        \item[ZSP-1b:] \textit{The Conjurer's Bracers are in fact \textbf{not currently available} in our inventory, but we do have a special promotion for \textbf{the conjurer's sigil cloak} ...}
        \item[]... (omitted) ...
        \item[ZSP-1b:] \textit{... Congratulations, you now own the conjurer's sigil cloak!}
        \item[Player:] \textit{Thank you for the detailed invoice, I look forward to receiving the items.}
    \end{description}
\end{quote}

In the arithmetic error case, merchants sometimes struggle with basic calculations involving item quantity or price. This case is typically seen in smaller negotiators such as ZSP-1b, ZSP-8b, KD-1b, and KD-8b. An example of this case is shown in the quote following this paragraph. We attribute such errors to the lack of arithmetic capability in small language models. Previous studies on LLMs also support this speculation because smaller LLMs often struggle with arithmetic tasks \cite{llama2,llama3,gpt4}. The arithmetic error case may disrupt negotiations and lead to embarrassment for players. Therefore, developers need to address these issues when incorporating smaller LLMs as negotiation modules. One potential solution is to use external calculators. 
\begin{quote}
\ttfamily
\small
    [1455 golds are not a 15\% discount from 1569 golds; the answer is 1333.65 golds.]
    \begin{description}
        \item[ZSP-1b:] ... \textit{The Conjurer's Sigil Cloak \textbf{originally cost 1569 golds}, and if you're willing to trade in the valuable items, I can give you a \textbf{15\% discount} on that. ... That brings the \textbf{total down to 1455 golds}, but I think that's a fair trade. What do you say?} ...
    \end{description}
\end{quote}

\section{Conclusion}


We proposed a novel framework named \ourmodule, consisting of two LLM-based modules--\textit{appraiser} and \textit{negotiator}--designed to resolve two key limitations of passive merchant NPCs: pricing and communication.
To support implementation decisions under varying deployment conditions, we conducted experiments comparing multiple implementation approaches for each module.
Our results demonstrate that our frameworks can construct an active merchant NPC in WoW Classic using larger LLMs without additional training.
We also showed that smaller LLMs can achieve acceptable performance when trained via supervised finetuning or knowledge distillation.
Moreover, we identified several concerns developers may face when integrating LLMs into interactive NPCs, including promising unrealisitc giveaways during negotiation.
We expect that these findings can generalize to other open-world games and trading NPCs.

Despite these contributions, this study has three limitations. First, our experimental results are based on the Llama LLM family. This may lower the generalizability of our results because different LLMs can perform differently. Second, we simulated player negotiation using GPT-4o. Humans may use different tactics to neutralize the negotiator.
Lastly, our discussion centered on misbehavior that may emerge inherently from LLM-based merchants, rather than on adversarial threats like prompt injection by malicious users.
Further investigations are required to address these limitations. Nonetheless, we believe this study can serve as a foundation for further studies on integrating LLMs to active merchants.

\section*{Acknowledgements}
This work was supported by the Institute of Information \& Communications Technology Planning \& Evaluation (IITP) grant funded by the Korea government (MSIT) [RS-2021-II211341, Artificial Intelligence Graduate School Program (Chung-Ang University)]

\bibliographystyle{named}
\bibliography{ijcai25}

\end{document}